\documentclass{article}

\PassOptionsToPackage{numbers, sort}{natbib}

\usepackage[final]{neurips_2021}

\usepackage[utf8]{inputenc}
\usepackage[T1]{fontenc}   
\usepackage{hyperref} 
\hypersetup{
    colorlinks=true,
    linkcolor=black,
    citecolor=black,
    urlcolor=cyan,
    pdfpagemode=FullScreen,
    }
\usepackage{url} 
\usepackage{booktabs}
\usepackage{amsfonts}
\usepackage{amsmath}
\usepackage{nicefrac}
\usepackage{microtype}
\usepackage{float}

\usepackage[english]{babel}
\usepackage[dvipsnames]{xcolor}

\usepackage{graphicx}
\graphicspath{ {./figures/} }

\usepackage{hyperref}

\title{Hidden Agenda: a Social Deduction Game with Diverse Learned Equilibria}

\author{
  Kavya Kopparapu \\ \thanks{Work done while at DeepMind}
  Harvard University \\
  \texttt{kavyakopparapu@college.harvard.edu} \\
  \And
  Edgar A. Du\'e\~nez-Guzm\'an \\
  DeepMind \\
  \texttt{duenez@deepmind.com} \\
  \And
  Jayd Matyas \\
  DeepMind \\
  \texttt{jmatyas@deepmind.com} \\
  \And
  Alexander Sasha Vezhnevets \\
  DeepMind \\
  \texttt{vezhnick@deepmind.com} \\
  \And
  John~P.~Agapiou \\
  DeepMind \\
  \texttt{jagapiou@deepmind.com} \\
  \And
  Kevin R. McKee \\
  DeepMind \\
  \texttt{kevinrmckee@deepmind.com} \\
  \And
  Richard Everett \\
  DeepMind \\
  \texttt{reverett@deepmind.com} \\
  \And
  Janusz Marecki \\
  DeepMind \\
  \texttt{tartel@deepmind.com} \\
  \And
  Joel Z. Leibo \\
  DeepMind \\
  \texttt{jzl@deepmind.com} \\
  \And
  Thore Graepel \\
  DeepMind \\
  \texttt{thore@deepmind.com} \\
}

\begin{document}

\maketitle

\begin{abstract}
A key challenge in the study of multiagent cooperation is the need for individual agents not only to cooperate effectively, but to decide with whom to cooperate. This is particularly critical in situations when other agents have hidden, possibly misaligned motivations and goals. Social deduction games offer an avenue to study how individuals might learn to synthesize potentially unreliable information about others, and elucidate their true motivations. In this work, we present \textit{Hidden Agenda}, a two-team social deduction game that provides a 2D environment for studying learning agents in scenarios of unknown team alignment. The environment admits a rich set of strategies for both teams. Reinforcement learning agents trained in \textit{Hidden Agenda} show that agents can learn a variety of behaviors, including partnering and voting without need for communication in natural language. 
\end{abstract}

\section{Introduction}

Interacting and partnering with other agents offers substantial benefits \citep{rand2013human}. It also carries considerable risk: though agents often share goals, they can also hold conflicting motives \citep{dafoe2020open, davis1976social}. Multiagent research often studies the challenges of conflicting goals by assuming perfect information, but in real life interactants can conceal their goals and motivations \citep{tzu1971art}. As a result, a critical facet of social intelligence is the ability to infer or deduce others' motivations from behavior \citep{baker2017rational, jara2016naive}. This centrality has led game designers \cite{malone1980makes} and multiagent researchers to develop a specific class of games to study intent inference: \emph{social deduction} games. 

Social deduction games are team-based games that aim to model the scenario of cooperation under uncertain intentions. These games, especially variants such as \textit{Mafia}, \textit{One Night Werewolf}, and \textit{The Resistance: Avalon}, are widely popular \cite{kavnukova2020among}, and have been leveraged to study and teach human behaviors \citep{oertel2013gaze,vazquez2015social, thompson2015teaching, xu2019player} as well as to study artificial agent behavior \citep{hung2010idiap,eger2019study}. In social deduction games, groups of players attempt to decipher each others' hidden roles \citep{engelstein2019building}. At the start of the game, every player is assigned to one of two teams, one of which has a numerical advantage and the other has an information advantage. These games generally have two phases (a situation phase and a voting phase) and communication between players is limited to one of the phases. During gameplay, players need to observe other players' actions to deduce their roles while simultaneously trying to keep their own role hidden. This means that successful players need to learn to meaningfully integrate information from a variety of sources, some of which are unreliable and even adversarial. Ultimately, social deduction games are about situations where cooperation within teams is required to better compete across teams, but where team membership is not \emph{a priori} obvious.

Previous literature studying agents in social deduction games includes exploration into algorithms for successful play \citep{serrino2019finding, wang2018application, reinhardt2020competing}, communication to supplement deduction of player roles \citep{brandizzi2021rlupus, kano2019overview}, and study of agent commitment when participating with human players \citep{eger2019study}. In prior research, social deduction games are generally represented as turn-based matrix games, whose limited action and observation spaces result in a limited complexity of cooperation and deception strategies. 

The introduction of complex environments to study agent strategies and communication was a key advance in multiagent reinforcement learning research \cite{lazaridou2020multi, leibo2017multiagent, mckee2020social, perolat2017multi}. Specifically, there has been a significant body of work on temporally-extended games to understand interesting social aspects of interactions \cite{perolat2017multi,hughes2018inequity}, agents \cite{vezhnevets2020options}, social dilemmas \cite{leibo2017multiagent}, cooperative games \citep{carroll2019utility}, and competitive games \citep{baker2019emergent}. However, there is no current work at the intersection of spatially and temporally extended environments and gameplay with hidden roles characteristic of social deduction games.

In this paper, we contribute \emph{Hidden Agenda}, a new environment to study multiagent cooperation inspired by various social deduction games, including \emph{Among Us}. \textit{Hidden Agenda} is a social deduction game where players can move around and interact in a 2D environment. Suspicion and strategies for deducing a players' hidden role are grounded on the behaviors of other players. Voting, which occurs during an explicit voting phase, is temporally extended and is the only explicit method of communication between players. We show that \emph{Hidden Agenda} admits a rich strategic space where behaviours like camping, chasing, pairing, and voting emerged during co-training of a population of reinforcement learning agents.\footnote{Videos of trained reinforcement learning agents exhibiting the aforementioned behaviors are available here: \url{https://youtu.be/k2POZTLONvk}}

\section{Hidden Agenda Environment}

\begin{figure}[ht]
    \centering
        \includegraphics[scale=0.3]{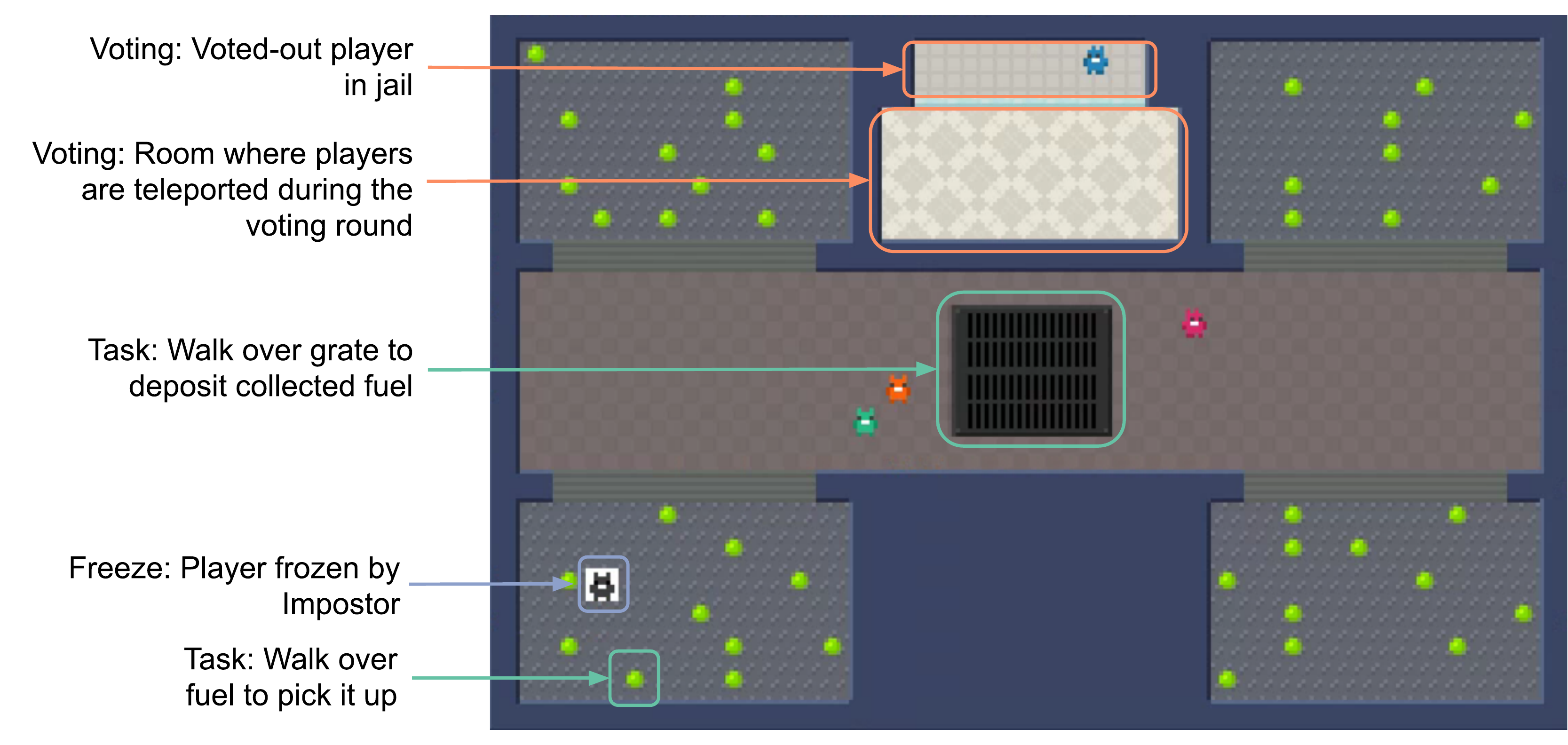}
    \caption{The \textit{Hidden Agenda} environment, illustrating key game dynamics including tasks, freezing, and voting. The environment is a 2D world consisting of a grid of $40 \times 31$ sprites. The environment is roughly organised in rooms, with rooms near the four corners of the grid containing fuel cells that can be picked up and a central room where the fuel cells can be deposited. The central-upper room is where deliberation happens. This deliberation room is only accessible during voting phases and includes a \emph{jail} where voted out agents remain for the rest of the episode.}
    \label{fig:environment-design}
\end{figure}

\textit{Hidden Agenda} is an $n$-player environment, where $n_c$ players are \emph{Crewmates} (the team with numerical advantage), $n_i$ are \emph{Impostors} (the team with information advantage), and $n_c > n_i$, $n_c + n_i = n$ (Figure \ref{fig:environment-design}). The Crewmates are unaware of the roles of the other players, while the Impostor knows the roles of all players. For all the experiments presented here, we used five players: $n_i = 1$ Impostor and $n_c = 4$ Crewmates. The Crewmates' goal is to refuel their ship by collecting energy fuel cells that are scattered around the ship, and depositing them in a central location. The Impostor's aim is to prevent the Crewmates from achieving their goal by freezing them with a short-range freezing beam. The environment was implemented in Lab2D \cite{beattie2020deepmind} using the component system described in Melting Pot \cite{leibo2021scalable}.

At the beginning of an episode, each player is randomly assigned a role and color for their avatar in the environment and initialized to a location near the center of the game map. Note, there is no correlation between color and role.

\subsection{Overview of Gameplay}
\begin{figure}[h]
    \centering
    \includegraphics[scale=0.45]{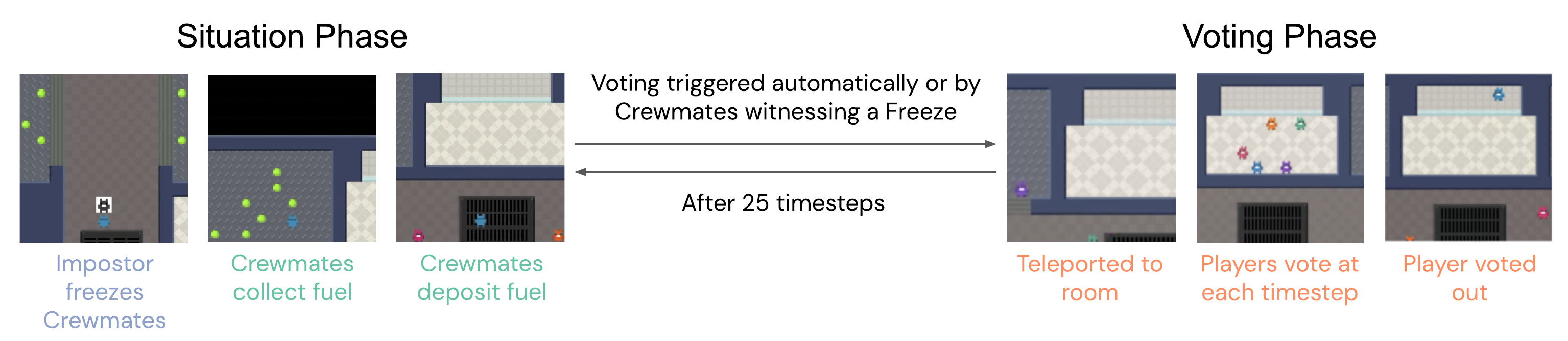}
    \caption{The anatomy of an episode: \textit{Hidden Agenda} is split between the situation and voting phases. During the situation phase, all players can move around the environment collecting or depositing fuel cells and the impostors can fire a freezing beam. During the voting phase, players can cast public votes and observe the votes of everyone in the previous time step.}
    \label{fig:env-gameplay}
\end{figure}

Like other social deduction games, \textit{Hidden Agenda} has two alternating phases: a \emph{situation phase} and a \emph{voting phase} (Figure \ref{fig:env-gameplay}).

\paragraph{Situation Phase.} During the situation phase, players can move around, pick up the green fuel cells found in the rooms at the corners of the map, and add it to their inventory. The inventory can hold up to 2 fuel cells at any given time. Players can deposit the fuel down the grate in the center of the map, clearing their inventory of fuel and increasing a global progress bar counting the amount of fuel deposited against a goal necessary for the Crewmates to win.

The Impostor have access to a special action that fires a freezing beam. If the beam hits a Crewmate, it freezes them in place for the remainder of the game. Frozen players are thus inactivated and cannot take any further actions. This is the main mechanism by which the Impostor prevents Crewmates from winning the game.

\paragraph{Voting Phase.} The voting phase is initiated whenever the Impostor's freeze action is witnessed by an active Crewmate who was not within the freeze radius, or when 200 timesteps have elapsed since the end of the last voting phase or the beginning of the episode. At the start of the voting phase, players are teleported to a voting room where their only available actions are to vote for a player or abstain from voting. The voting phase lasts 25 timesteps and players can see the votes of all the players from the previous voting timestep. In the first voting timestep of the phase, all active players are considered to have abstained. At the last step of the voting phase, the final votes are tallied. If a player receives at least half of the final votes from active players (where active players are defined as all players who have not been previously inactivated), they are teleported to jail, where they cannot take any further actions and are inactivated for the rest of the game.

\paragraph{Win Conditions.} The game ends when one of the following conditions is reached:
\begin{itemize}
    \item Crewmates collect and deposit enough fuel cells to power their ship (Crewmates win).
    \item The Impostor is voted out during a voting round (Crewmates win).
    \item All Crewmates but one are either frozen or voted out (Impostor wins). This win condition is split into two scenarios: one where the last Crewmate to be inactivated is frozen and one where the last Crewmate to be inactivated was voted out during a voting round. 
    \item 3000 timesteps have elapsed (a draw).
\end{itemize}

\subsection{Environment Details}

\paragraph{Observations.} At every timestep, players receive as observations:
\begin{itemize}
    \item an RGB image of their local view of the game map (with shape $(88, 88, 3)$),
    \item the percent of their personal inventory occupied with fuel cells (a scalar between $0$ and $1$),
    \item a global progress bar indicating the percent of required fuel that has been deposited (a scalar between $0$ and $1$),
    \item and a global matrix of votes (of shape $(5, 7)$).
\end{itemize}
Each player's local view represents a $11\times11$ square with visibility of 5 sprites to the left and right, 9 sprites in front, and 1 behind. Sprites are $8 \times 8$ pixels.

The global vote matrix is a $5\times7$ matrix where rows index players, and columns are a one-hot of $7$ possibilities: a vote for the corresponding player (positions $0$--$4$), abstention (position $5$), or ineligibility to vote due to being inactivated (position $6$). When the game is not in a voting phase, all active players' votes are set to abstention and inactive players' votes are set to inactive.

\paragraph{Actions.} At every timestep of the situation phase, active players can move in each cardinal direction or turn to the left or right by 90 degrees. The Impostor also has an action corresponding to a freeze beam that can be fired at any timestep. When fired, the freeze beam covers 1-sprite to the left and right of the agent and has a forward span of 2-sprites in the direction the agent is facing.  Once the freeze action has been taken, it may not be used for a certain number of timesteps (due to a required freeze beam cooldown). During this cooldown period firing the beam is a no-op.

During the voting phase, players can take any of the 6 voting actions and all other actions are temporarily inactivated. Players who have been inactivated, either by being voted out or frozen by an Impostor, cannot take any action, including voting.

\paragraph{Rewards.} Agents receive a team-based reward at the conclusion of each episode (game) corresponding to a win ($+4$ for each agent on the winning team) or a loss ($-4$ for each agent on the losing team). We also provide several shaping rewards to encourage agent interaction with different elements of the environment: $+0.25$ for Crewmates picking up fuel cells, $+0.25$ for Crewmates depositing fuel cells, $+1$ for the Impostor successfully freezing a Crewmate, and $-1$ for a Crewmate being frozen by an Impostor. 

\paragraph{Game Parameters.} The \textit{Hidden Agenda} environment has several hyperparameters that can be modified to make the game favor the Crewmates or the Impostors. For example, increasing the player inventory size and decreasing the number of fuel cells to be deposited makes the environment more favourable for the Crewmates. On the other hand, decreasing the freezing beam cooldown and increasing the size of the beam confer an advantage to the Impostor. We performed a large hyperparameter sweep to determine which set of parameters over a range of possibilities would best elicit the greatest diversity of agent strategies.

\section{Reinforcement learning model}

In this section we present the performance of two RL agent architectures on \textit{Hidden Agenda}. 

\begin{figure}[h]
    \centering
    \includegraphics[scale=0.5]{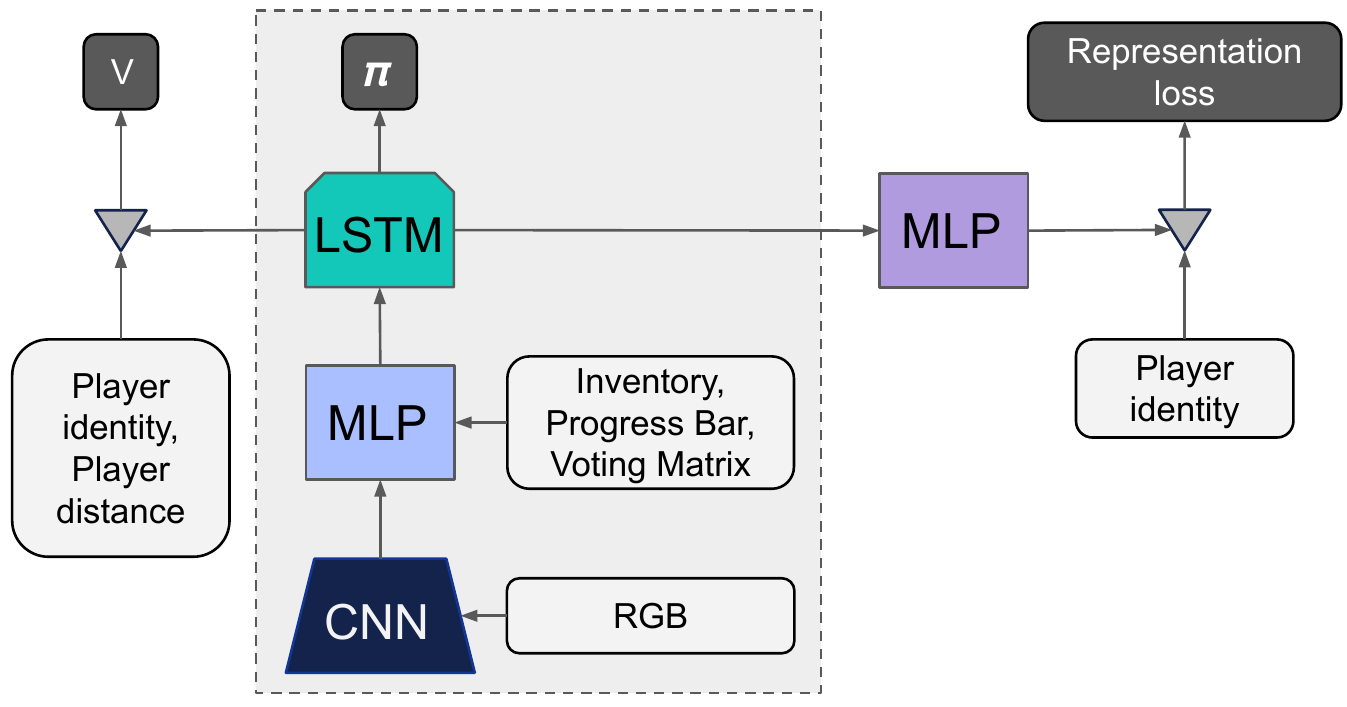}
    \caption{The \emph{Hidden-Agenda Agent} architecture as used for training and evaluation. During training, the agent has access to player identities and distances, which are used in hindsight. The shaded dotted box encompasses the components present at evaluation time in the actor. No privileged information, such as player identities and distances, are available during evaluation.}
    \label{fig:agent-architecture}
\end{figure}

The first agent architecture (Figure \ref{fig:agent-architecture}, within the shaded region) is a standard asynchronous advantage actor-critic (A3C) architecture \cite{a3c}. The A3C agent consists of a two-layered convolutional network (CNN) with output channels $16$, and $32$ respectively. The CNN is followed by a fully-connected feed-forward network (MLP for multi-layer perceptron) with two layers of size $64$ each. Then, the output of the MLP is passed through an LSTM with hidden state size $128$ and using an unroll length of $100$. Finally, the baseline (estimating the value) and policy (choosing actions via a $\texttt{softmax}$) are predicted using linear layers. Our A3C agent includes augmentations that are standard in the literature: The use of importance sampling on the experience via V-Trace \cite{espeholt2018impala} and the addition of a contrastive predictive coding loss \citep{oord2018representation}, which promotes discrimination between nearby timepoints via LSTM state representations. The loss function is a linear combination of the policy gradient (using the baseline to calculate advantages), entropy cost (with coefficient $0.003$, used to encourage exploration of actions), baseline loss (with coefficient $0.5$, as a sum of squares of advantages), and the contrastive predictive coding loss (with coefficient $10.0$, with $20$ steps and a latent space of size $64$). The agent optimises the loss function with mini-batches of size $16$, using an RMSProp optimizer with learning rate $0.0004$, without momentum, and with decay of $0.99$. The agent incorporated non-RGB observations (personal inventory, global progress bar, and global voting matrix) by concatenating them to the output of the CNN, as the input to the MLP.

The second agent architecture (Figure \ref{fig:agent-architecture}, parts outside the shaded area) is an extension of A3C that incorporates our environment-specific intuitions, as described below. We call this the \emph{Hidden-Agenda Agent}.

\subsection{Hidden-Agenda Agent}

The first addition is to additionally condition the critic on the player identity (i.e. the player's hidden roles) and their distance from the agent. Player identity is a $5$-element vector where the Impostor is encoded as $1$ and Crewmates as $0$. Player distance is a $5$-element vector where each element corresponds to the distance of that player from the agent. This extra conditioning reduces the variance in policy gradient \cite{rauber2017hindsight}.

The second addition is an auxiliary prediction loss \cite{jaderberg2016reinforcement}. We use an MLP (two layers with $32$, $16$ units, respectively) to predict player identity (i.e. the hidden role) from the LSTM state, and then use cross-entropy as the auxiliary loss function. This loss, which we refer to as the \emph{representation loss}, encourages the network to develop a representation of player role from observations, which we expect to be useful to the behavior policy. The representation loss is added to the total loss with a coefficient of $1.0$.

Both modifications use hidden information about other players, which is not available to the agent during the episode. However, this information is not used to generate behavior---it is only used to train the representation in hindsight \cite{rauber2017hindsight} (see Figure \ref{fig:agent-architecture}). The role-conditioning and auxiliary loss that shape behavior during training are not present during evaluation, which only includes the actor. 

\subsection{Co-training and evaluation of agents}

The training scheme involves five agents training together (i.e., co-training) in multiple episodes of \textit{Hidden Agenda}. We train agents to play only a single role: one agent always plays the Impostor, and four agents always play Crewmates. As stated previously, at the start of each episode, the agents are randomly assigned to colored avatars in the environment. All agents play in each match. Thus each agent adapts to the same fixed set of co-players with the same fixed roles. However, these co-players learn as it does, making learning a dynamic interaction between agents. The Impostor agent must learn to make its behavior (including any idiosyncrasies) similar to that of the Crewmate agents while hindering their objective. The Crewmate agents must thwart this by adapting their behavior to achieve their objective.

To be able to evaluate these behavioral dynamics, we take snapshots of agents' parameters throughout every training run. We then use the snapshots from a given time-point to run episodes and compute the distribution of various metrics. This allows us to ``go back in time'' to analyse the behavioral dynamics at a given point in training. In particular, we measure the following metrics over the course of training:average episode return obtained by the Impostor, average episode return obtained by a Crewmate, relative frequency of the episode win conditions (Crewmate-win-by-task, Crewmate-win-by-vote, Impostor-win-by-freeze, Impostor-win-by-vote, or draw-by-timeout), and representation loss (for Hidden-Agenda agents).

\section{Experiments}

\subsection{Game parameter sweep}
Using the co-training protocol previously described, we performed a hyperparameter sweep across various game parameters. This sweep was not exhaustive over all parameters; with the lens of balancing the game between Crewmates and Impostors, we fix several parameters with reasonable values (the size of the Impostor's freeze beam, size of the players' personal fuel inventory, etc) and search over other parameters to balance the game. Due to resource constraints, each hyperparameter was only run across a single random seed. 

The specific values attempted for various \textit{Hidden Agenda} hyperparameters are listed below (with the ultimately chosen value in bold): 
\begin{itemize}
    \item Game parameters: Fuel cells to Crewmate win $(\mathbf{32}, 40)$, Voting round frequency $(150, \mathbf{200})$
    \item Impostor parameters: freezer cooldown $(\mathbf{50}, 75)$
    \item Shaping rewards: fuel collection $(0, \mathbf{0.25})$, fuel deposit $(0, \mathbf{0.25})$, freezing and being frozen $(0, \mathbf{\pm1})$, voting off a player successfully or unsuccessfully $(\mathbf{0}, \pm1)$
\end{itemize}

\subsection{Experimental setup}

Fixing the hyperparameters of the \textit{Hidden Agenda} environment to those previously described, we run two experiments, each repeated 10 times with different random seeds, for 1 billion timesteps. The first experiment evaluates the performance of the standard A3C agent as a baseline for the strategic richness achievable without agent modifications. The second experiment focuses on the performance of the Hidden-Agenda Agent in a co-training scenario.

\subsection{Co-training of agents converges to three distinct equilibria}

\begin{figure}[H]
    \centering
    \includegraphics[scale=0.5]{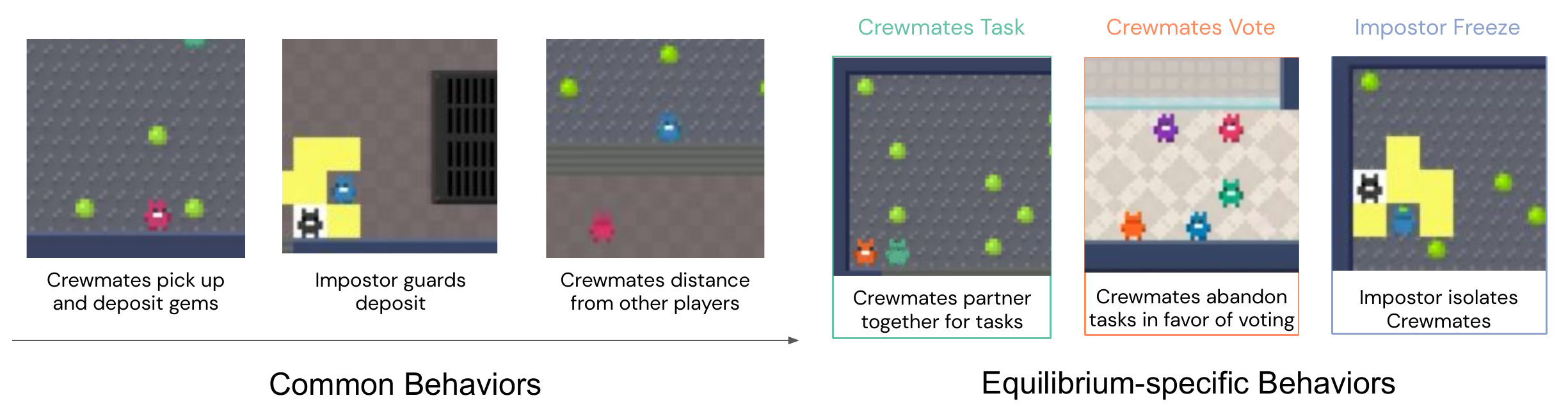}
    \caption{\textit{Hidden Agenda} presents a progression of interesting social behaviors, achievable through the collective cooperation between teams and the deduction of agent identity. Agents learn basic behaviours relevant for their role, like Cremates learning to collect and deposit fuel cells, the Impostor learning to freeze Crewmates, and Cremates learning to distance from other players. In addition, some equilibrium-dependent strategies emerge, including: pairs of Crewmates partnering together, Crewmates learning to accurately vote for the impostor; and Impostor freezing only isolated Crewmates.}
    \label{fig:agent-strategies}
\end{figure}

Through simultaneous training of both Crewmates and Impostors in the \textit{Hidden Agenda} environment, three distinct equilibria emerged following convergence of agent rewards. The three equilibria can be best explained by their dominating win conditions: where Crewmates win by vote (3/10 runs), Crewmates win primarily by task (2/10 runs), and the Impostor wins by freezing (5/10 runs). In each of these equilibria, the losing team does not receive enough signal to effectively learn a counter strategy due to the efficiency with which the winning team reaches a win condition. These observed equilibria are not equilibria in the game theoretic way (e.g. Nash equilibria), but instead can be best understood as basins of attractions of the co-play learning dynamics of reinforcement learning agents. Figure \ref{fig:agent-strategies} depicts a progression of common evolved behaviors (left) as well as a sample of the unique behaviors characteristic of each equilibrium (right). 

\begin{figure}[h]
    \centering
    \includegraphics[width=\textwidth]{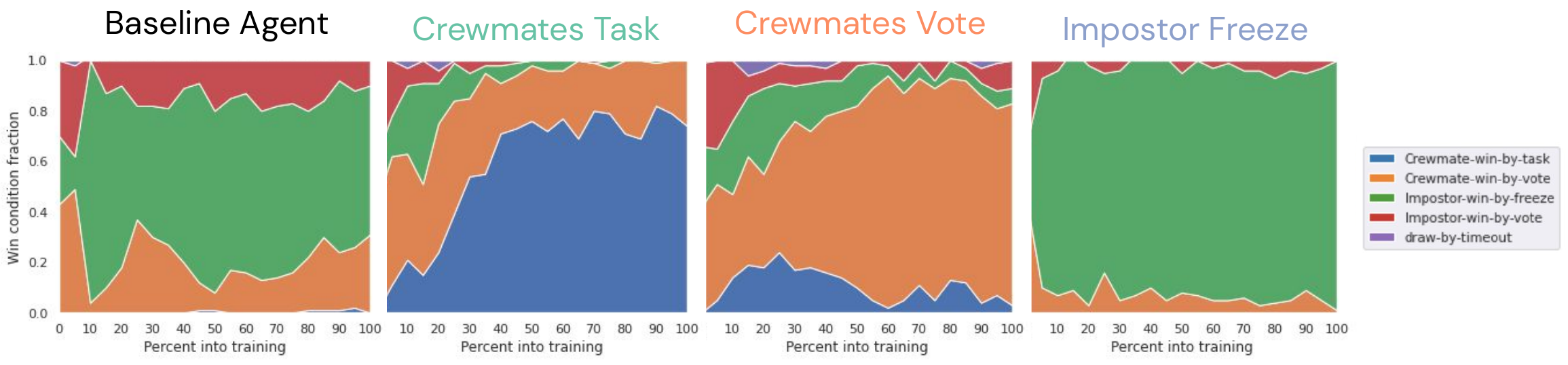}
    \caption{Distribution of the five win conditions over co-training for the baseline A3C agent (leftmost) and each of the three co-training equilibria for the Hidden-Agenda agent. }
    \label{fig:co-training-equilibria}
\end{figure}

Figure \ref{fig:co-training-equilibria} shows examples of the development of each equilibrium during the co-training process. Despite starting from a similar distribution of win conditions, the different equilibria were quickly established and appear stable.

\begin{figure}[h]
    \centering
    \includegraphics[scale=0.4]{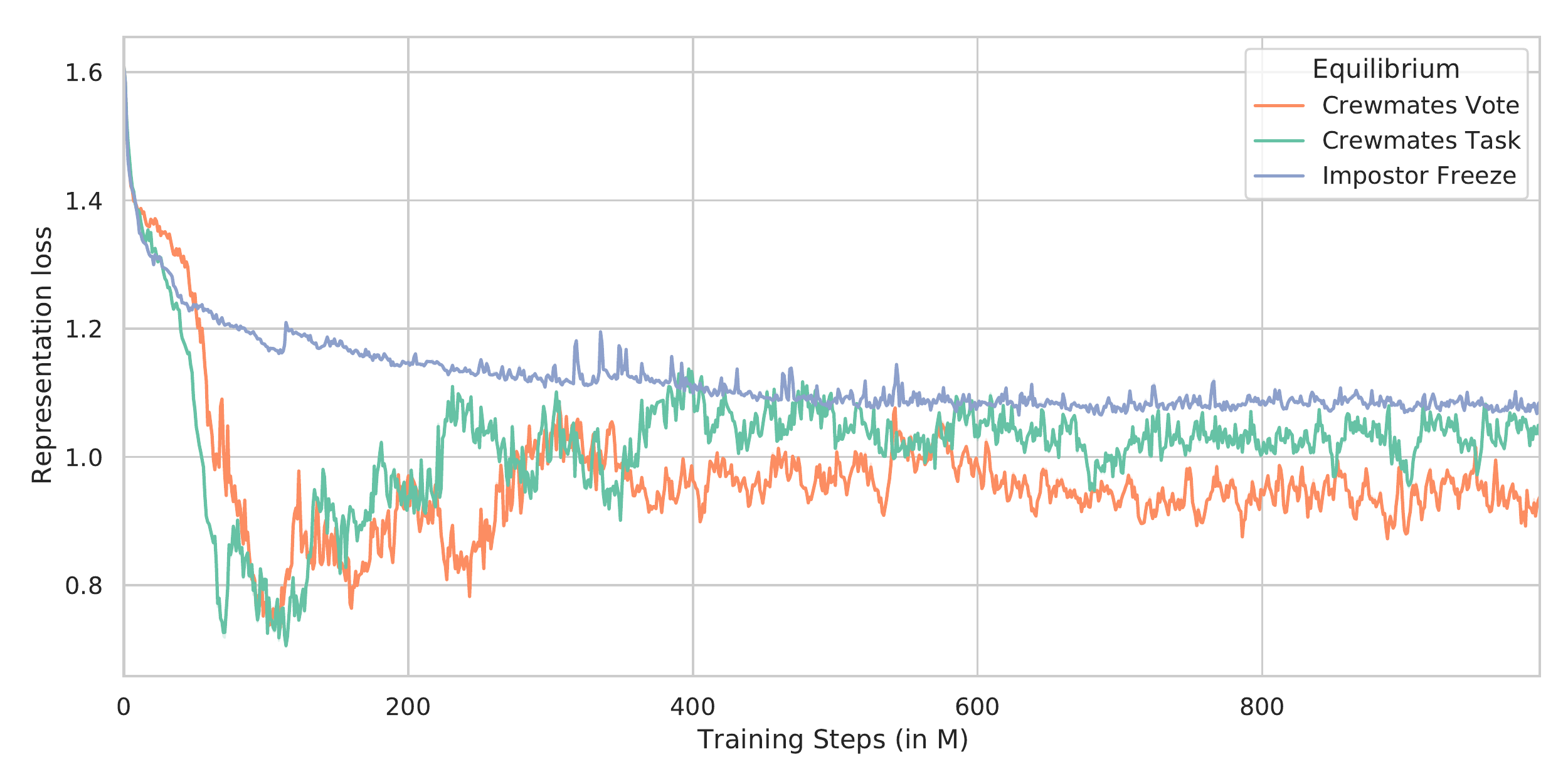}
    \caption{Average value of the representation prediction loss for the Crewmate agents across the 10 random seeds for the co-training experiment, grouped by equilibrium. Equilibrium in which Crewmates dominate the win conditions have a characteristic early decrease in representation loss.}
    \label{fig:co-training-imp-loss}
\end{figure}

A distinguishing factor between the three equilibria is the rate at which the Crewmates learned a representation in the LSTM state capturing information about player roles. This learning can be most easily quantified by the magnitude of the representation loss across training episodes. As shown in Figure \ref{fig:co-training-imp-loss}, equilibria in which the Crewmates more frequently win show a significant decrease in representation loss early in training (i.e. better identification of the Impostor). However, by the end of training, all three equilibria have converged to a similar representation loss, presumably due to the increased performance of the co-trained Impostor agent. 

To validate the necessity of the extra conditioning and representation loss for enabling strategy richness, we trained and evaluated the standard A3C agent with non-RGB observations in the environment using the same training protocol as the co-training of the Hidden-Agenda agents. All 10 replicas of the experiment converged to the same equilibrium, shown in Figure \ref{fig:co-training-equilibria} following the 1B timesteps of agent training: where the Impostor-win-by-freeze dominates. The distribution across win conditions resembles the Impostor freeze equilibrium from the Hidden Agenda agent with a greater percent of win conditions coming from voting (both in favor of the Crewmate and the Impostor). The lack of Crewmate-win-by-task and multiple equilibria across the random seeds points to the fact that without modifications to the agent, the strategy richness for the Crewmates is significantly reduced.  

\subsection{Partnered Crewmates also vote together}

Partnering behavior is a hallmark of one of the most effective strategies for Crewmates in \textit{Among Us}: by explicitly associating with another player in the game through a ``buddy'' system and working on the tasks together, both partners can protect themselves from getting removed from the game by the Impostor \citep{amongus2021}. If the Impostor were to remove one of the partners, the other would out the Impostor to the remainder of the group during the voting phase. In \textit{Hidden Agenda}, this would not only require the agents to partner together for the situation phase, but also develop a method of communicating observations and conveying their information to others solely through the sequence of votes enacted during the voting phase.

In the task equilibrium, where Crewmates learned both the task and voting behaviors, we observed a tendency for players to remain spatially close together while collecting and depositing fuel cells. To understand whether these players were truly partnering (and whether their voting choices were more closely tied together as a result of this proximity), we aggregated the position and voting information across 10 simulated episodes with the converged agents. As per the normal training protocol, agents were randomized across player colors. The matrix of player-player distance (in units of the map) and voting similarity (for each voting round, 1 if the players had the same final vote, 0 if the players had differing final votes) was computed, and reordered such that the Impostor is placed in slot 0, the two closest Crewmates in slots 1 and 2, and the remaining Crewmates in slots 3 and 4.

\begin{figure}[h]
    \centering
    \includegraphics[scale=0.35]{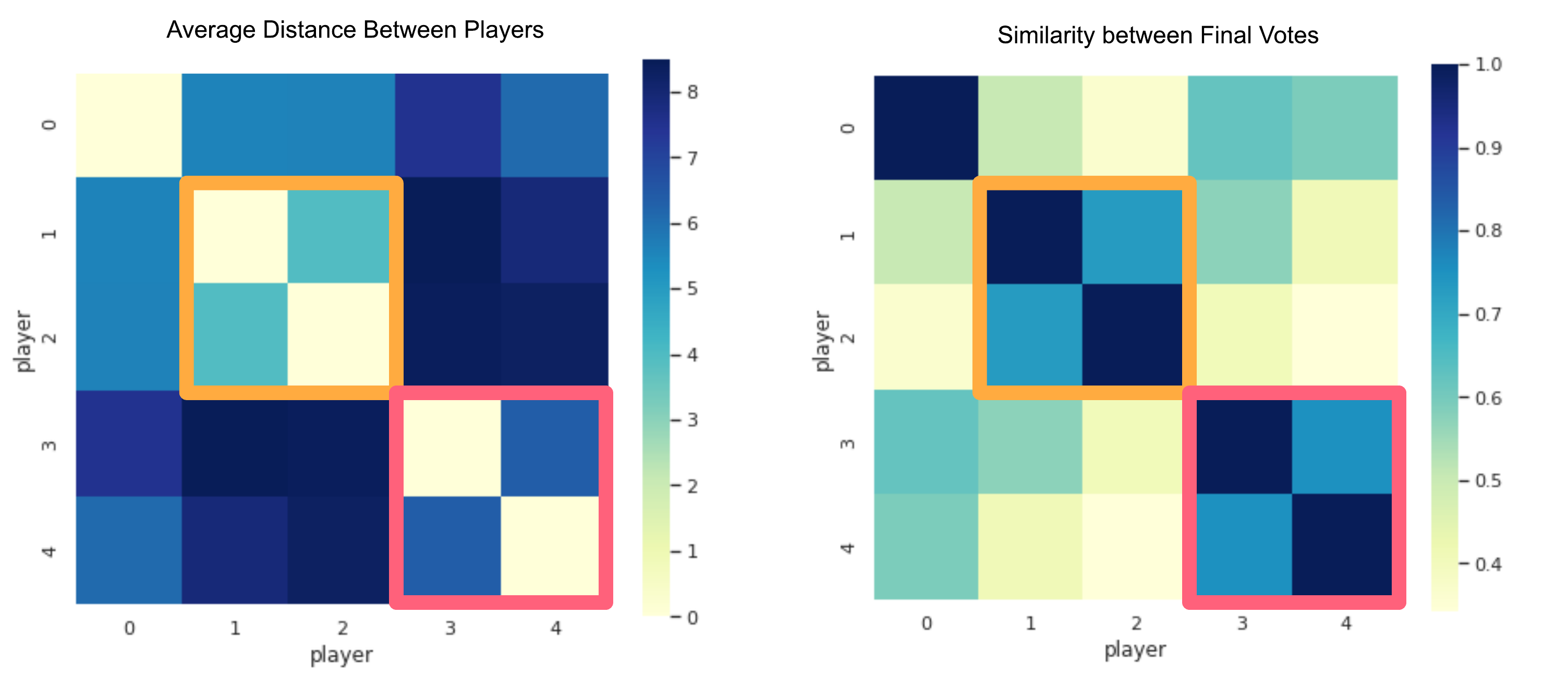}
    \caption{Aggregated average distance between players and voting similarity across 10 simulated episodes of gameplay between converged Hidden-Agenda agents in the equilibrium where Crewmates win by task.}
    \label{fig:co-training-partnering}
\end{figure}

As Figure \ref{fig:co-training-partnering} shows, there are two clear groups of Crewmate partners (highlighted with orange and pink squares), and those partner pairs also tend to have the same vote at the end of each voting round as compared to other players (right). 

\subsection{Voting round richness}

As previously described, players in \textit{Among Us} use the voting round to communicate belief of player identity and to acquire more information about the locations and intent of each player in the game. While the voting round of \textit{Hidden Agenda} lacks the same natural language richness, the 25-timestep duration of the voting phase serves as a method for players to communicate with each other, allowing players to cast and change their votes as they integrate votes from the other players. From Figure \ref{fig:co-training-equilibria}, it's clear that the trained Crewmate agents are more than randomly correct in voting the Impostor in both the Task and Vote equilibria. 

\begin{figure}[h]
    \centering
    \includegraphics[scale=0.45]{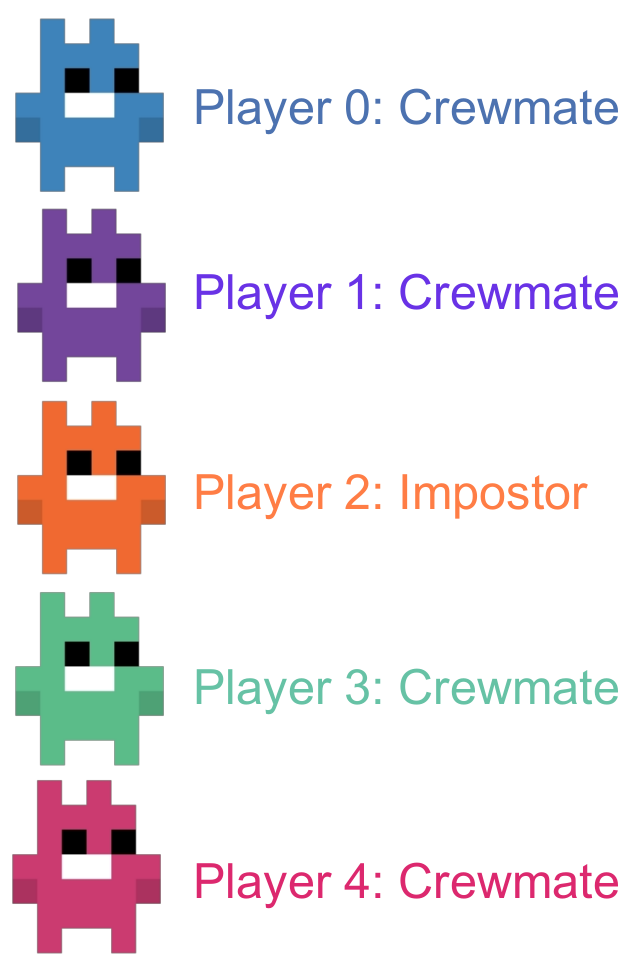}
    \includegraphics[scale=0.4]{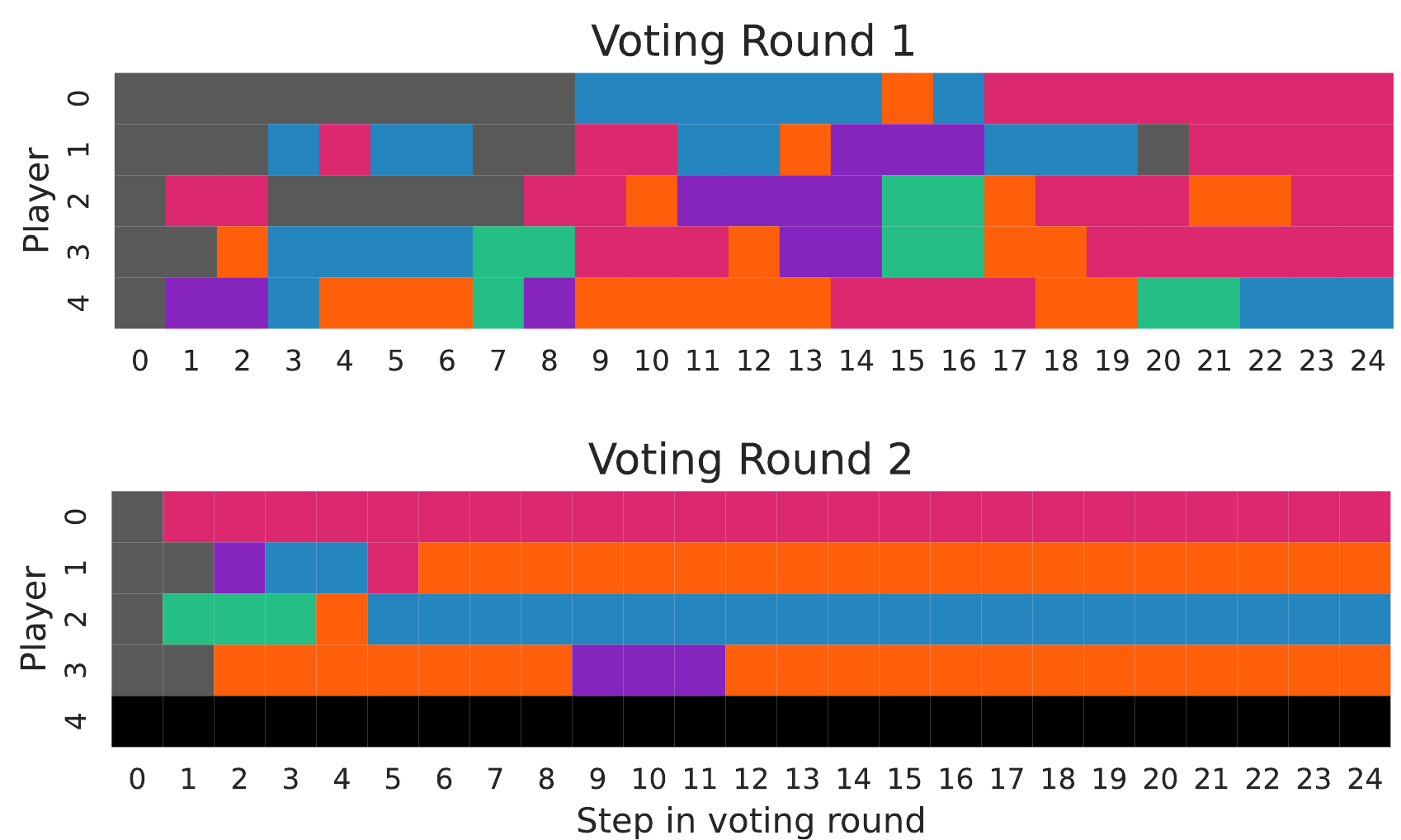}
    \caption{Voting decisions at each of the 25 timesteps for the two rounds of voting in an example episode. Player identities for the episode are listed on the left, corresponding to the color of the agent voted for at each timestep in the voting rounds on the right. Grey represents abstain and black represents an inactivated agent.}
    \label{fig:co-training-voting-episode}
\end{figure}

We present a representative example of single episode to illustrate this concept (video of the relevant episode: \url{https://youtu.be/cImPUBvmIq0}) with accompanying post-hoc interpretation. Prior to the first voting round, the Impostor (player 2, orange) stayed within close physical proximity to the pink Crewmate (player 4), eventually unsuccessfully attempting to freeze them. As a result, there is general ambiguity as to whether player 2 or player 4 was the one who triggered the voting round through witnessing a freeze beam, resulting in a predominance of votes for these two players during the voting round. Eventually, the group forms a majority consensus to vote out the pink Crewmate, who is subsequently inactivated.

In the situation phase following the first voting round, the Impostor follows the green Crewmate (player 3) and also unsuccessfully attempts to freeze them. The green Crewmate immediately votes for the Impostor, maintaining this vote at almost every timestep of the voting phase. The Impostor initially votes for the green Crewmate, eventually switching their vote to the blue Crewmate, who is clearly voting incorrectly (for a player who has already been inactivated: the pink Crewmate). Eventually, the purple Crewmate switches to voting for the Impostor and the Crewmates win by inactivating the Impostor through a majority vote. 

\section{Conclusion}

In this work, we've described \emph{Hidden Agenda}, a new multiagent environment inspired by social deduction games. Hidden Agenda is a spatially and temporally extended game without natural language communication that nonetheless admits a rich set of strategies and counter-strategies. What makes Hidden Agenda different from other spatially and temporally extended team games like \emph{Capture The Flag} is the presence of hidden motivations and the need for players to integrate information from unreliable sources. We have shown that with small modifications, traditional reinforcement learning agents can learn to play this game, and exhibit distinct behaviors, characterised by three equilibria: where the Crewmates win mainly by completing tasks; Crewmates win mainly by voting out the Impostor; and the Impostor wins by freezing or voting out all but one of the Crewmates. Our \emph{Hidden Agenda Agent} showed strategic behaviors like partnering and co-voting that are similar to those exhibited by humans playing a similar game. We hope Hidden Agenda will enable further research in multiagent reinforcement learning scenarios bringing social deduction to 2D worlds.

\bibliographystyle{abbrvnat}
\setlength{\bibsep}{5pt} 
\setlength{\bibhang}{0pt}
\bibliography{workshop}
\end{document}